\documentclass[11pt]{article}

\usepackage[hyperfootnotes=false]{hyperref}
\usepackage[margin=1in]{geometry}
\usepackage{graphicx}
\usepackage{float}
\usepackage{booktabs}

\usepackage{amsmath}
\usepackage{amssymb}
\usepackage{amsthm}
\usepackage{bm}

\usepackage{tikz}
\usepackage{lmodern}
\usepackage{xcolor}
\usetikzlibrary{
  arrows.meta,
  backgrounds,
  calc,
  fit,
  positioning,
  shapes.geometric
}
\definecolor{panel}{HTML}{F7F9FA}
\definecolor{shadow}{HTML}{E8ECEF}
\definecolor{ink}{HTML}{24313A}
\definecolor{muted}{HTML}{667680}
\definecolor{rule}{HTML}{B8C2C9}
\definecolor{paper}{HTML}{FFFFFF}
\definecolor{panel}{HTML}{F7F9FA}
\definecolor{globalLine}{HTML}{B14A9B}
\definecolor{globalFill}{HTML}{F7E3F4}
\definecolor{localLine}{HTML}{3975AD}
\definecolor{localFill}{HTML}{E1ECFA}
\definecolor{mfeLine}{HTML}{7356A7}
\definecolor{mfeFill}{HTML}{F0EAF7}
\definecolor{gradLine}{HTML}{2A7F62}
\definecolor{gradFill}{HTML}{E3F2EC}

\newtheorem{theorem}{Theorem}

\newcommand{\R}{\mathbb{R}}

\title{MENO: Memory-Efficient Neural Operator}

\author{Shengyang Xu\textsuperscript{a,1}, Weijun Zhang\textsuperscript{a,1}, Jun Hu\textsuperscript{a,c}, Pengzhan Jin\textsuperscript{b,c,*}}

\date{}

\begin{document}

\maketitle

{\renewcommand{\thefootnote}{a}\footnotetext{\fontsize{8.91}{10}\selectfont School of Mathematical Sciences, Peking University, Beijing 100871, China}}

{\renewcommand{\thefootnote}{b}\footnotetext{\fontsize{8.91}{10}\selectfont National Engineering Laboratory for Big Data Analysis and Applications, Peking University, Beijing 100871, China}}

{\renewcommand{\thefootnote}{c}\footnotetext{\fontsize{8.91}{10}\selectfont Chongqing Research Institute of Big Data, Peking University, Chongqing 401329, China}}

{\renewcommand{\thefootnote}{1}\footnotetext{\fontsize{8.91}{10}\selectfont These authors contributed equally to this work.}}

{\renewcommand{\thefootnote}{*}\footnotetext{\fontsize{8.91}{10}\selectfont Corresponding author. Email: jpz@pku.edu.cn}}

\begin{abstract}
We propose the Memory-Efficient Neural Operator (MENO) as a high-performance PDE neural solver based on the Manifold Function Encoder
(MFE). MENO features three primary advantages: (1) MENO has a significantly smaller memory footprint and much faster training speed than other popular architectures, with the memory footprint being independent of the data resolution, and therefore holds the potential for scaling up to large-scale models. (2) MENO can accept PDE inputs of arbitrary form, including arbitrary geometric domains and arbitrary discretizations. In particular, it is capable of handling cross-geometry scenarios, i.e., where the input functions and the output solutions are defined on different manifolds. (3) MENO exhibits strong generalization capability, and achieves the best accuracy on most of the benchmarks we tested, compared with the results reported in the literature. The code is available on GitHub at https://github.com/jpzxshi/MENO, and all numerical examples in this paper can be run with a single command to reproduce the reported results.
\end{abstract}

\section{Introduction}

Neural operators offer a data-driven approach to approximating mappings between function spaces, enabling efficient surrogate modeling of families of partial differential equations (PDEs). Representative architectures, including DeepONet \cite{lu2019deeponet,lu2021learning} and the Fourier neural operator (FNO) \cite{li2020fourier}, learn solution operators rather than individual solutions \cite{yu2018deep,raissi2019physics}. Subsequent developments have explored alternative operator representations, including the extension of DeepONet and FNO \cite{jin2022mionet,rahman2022u}, wavelet and multiwavelet formulations \cite{gupta2021multiwavelet,tripura2023wavelet}, attention-based approximation \cite{cao2021choose}, and convolutional architectures \cite{he2023mgno,raonic2023convolutional}. Beyond architectural developments, neural operators have also motivated new training and numerical solution paradigms. Physics-informed DeepONets \cite{wang2021learning} incorporate governing equations into operator learning to reduce reliance on paired solution data. Rather than using neural operators solely as standalone surrogates, hybrid iterative methods integrate them into classical numerical solvers \cite{hu2026hybrid,zhang2022hybrid}. In many applications such as the practical scenarios of hybrid methods, however, the computational domain varies together with the coefficients, forcing terms, and boundary conditions. This introduces an additional challenge: the PDEs may be defined on different domains across problem instances.

The second phase in the development of neural operators focuses on adaptability to PDEs on varying domains. Preliminary attempts have been made to handle PDE problems on varying domains via transfer learning \cite{goswami2022deep}, however, this approach still requires a certain degree of retraining. Practical operator learning methods for handling varying-domain problems mainly fall into two categories: deformation and extension. The main idea of the deformation strategy is to deform a large class of topologically or diffeomorphically equivalent domains into the same canonical domain, and then learn the operator mapping on this canonical domain, such as Geo-FNO \cite{li2022fourier}, DIMON \cite{yin2024dimon}, and D2D\&D2E \cite{xiao2026deformation}. In particular, \cite{xiao2026deformation} presents a comprehensive theoretical framework of this class of methods. Although the deformation strategy performs well in specific cases, it faces a limitation, i.e., it cannot handle domains that are not topologically homeomorphic. Extension-based operator learning methods overcome this limitation. The idea behind the extension strategy is very simple: we extend different domains, along with the functions defined on them, to a larger canonical domain. This is no longer constrained by the topological structure of the domains. There are many extension-based methods, representative examples such as DAFNO \cite{liu2023domain} with the characteristic functions, GINO \cite{li2023geometry} with the signed distance functions (SDF), MFE-MIONet \cite{hu2025manifold} with the dual representation. The graph- or point cloud-based architectures such as Graph U-Net \cite{gao2019graph}, MeshGraphNet \cite{pfaff2020learning}, UPT \cite{alkin2024universal}, AB-UPT \cite{alkin2025ab}, GAOT \cite{wen2026geometry}, can also be regarded as dual representation-based methods, with local integration encoders or processing layers. All these representations can be viewed as different ways of embedding the manifold function space into a large Banach space. Importantly, different embedding approaches can lead to completely different encoding performance and training efficiency, and can also severely affect the GPU memory overhead. This is also a key focus of the next stage of neural operator research.

To handle operator learning problems in real-world three-dimensional industrial scenarios, the training cost of varying-domain neural operators must be kept under control. The direct extension methods including the characteristic functions and the SDF, suffer from a severe efficiency bottleneck in three-dimensional problems, as their computation involves physical points throughout the entire three-dimensional space. By contrast, the dual representation is advantageous, because the encoder based on it only involves the manifold itself and the functions defined on it in computation. The Transolver series \cite{luo2025transolver++,wu2024transolver,zhou2026transolver}, the PCNO \cite{zeng2025point}, the MFE-MIONet \cite{hu2025manifold}, etc. all belong to this category, with global integration encoder. The encoders in these works all expand manifold functions in the dual space, obtaining the latent codes by computing the integrals of some global basis functions over the manifold. Nevertheless, there remains a fundamental difference in efficiency between them. For Transolver and PCNO, each of their layers requires integration operations, so the computational cost during training grows with data resolution and leads to severe GPU memory bottlenecks for large-scale problems. Although Transolver-3 reduces the peak GPU memory usage to within an acceptable range for a single GPU through certain techniques, the computational cost remains enormous. The MIONet \cite{jin2022mionet} as an integration-free architecture, combined with the MFE \cite{hu2025manifold}, exactly resolves the memory and computation bottleneck, however, its generalization capability still lags behind integration-based architectures. Therefore, we are strongly motivated to propose a new neural operator architecture that is integration-free while achieving generalization performance no worse than that of integration-based neural operators, and that maintains both computational efficiency and prediction accuracy.

\textbf{Contributions.} Our main contribution is proposing a novel neural operator architecture, the Memory-Efficient Neural Operator (MENO), with a remarkably small memory footprint independent of data resolution, as well as extremely fast training. MENO also has strong generalization ability, and it achieves the best accuracy on most benchmarks we tested. Moreover, MENO can handle arbitrary geometric inputs, including cross-geometry scenarios.

The remainder of this paper is organized as follows. We begin with introducing the Manifold Function Encoder (MFE) in Section \ref{sec:mfe}. In Section \ref{sec:architecture}, we formally propose the architecture of MENO in details. Section \ref{sec:experiments} presents several numerical experiments regarding to MENO. Finally Section \ref{sec:conclusion} summarizes this work.

\section{Manifold Function Encoder}
\label{sec:mfe}

We first briefly recall the Manifold Function Encoder (MFE) introduced in \cite{hu2025manifold}, as the cornerstone of MENO. MFE is able to efficiently encode the shape of the manifolds and the functions defined on them, with rigorous theoretical guarantees.

\textbf{Setting.} Consider the union of a finite number of \(k\)-dimensional Lipschitz manifolds contained in the $d$-dimensional box
\begin{equation}
    V:=[0,1]^d,
\end{equation}
where $0\leq k\leq d$.
More precisely, define
\begin{equation}
\begin{split}
    \mathcal{M}_k:=\{\cup_{i=1}^{m}M_i:&M_i\subset V{\rm\ is\ a\ compact\ }k{\rm -dimensional\ Lipschitz\ manifold\ }(\mathcal{H}^k(M_i)<\infty),\\ &M_i\cap M_j{\rm\ is\ manifold\ of\ dimension\ less\ than\ }k{\rm\ or\ empty\ set}, i\neq j,m\geq 1\},
\end{split}
\end{equation}
where \(\mathcal{H}^k\) is the \(k\)-dimensional Hausdorff measure, and
\begin{equation}
    \boxed{X_k:=\{(M,f_M):M\in\mathcal{M}_k,\ f_M\in L^2(M)\}.}
\end{equation}
Note that when \(k=0\), an element of \(\mathcal{M}_0\) is regarded as a finite set of points in \(V\), and \(f_M\in L^2(M)\) is a real-valued function on these points. $X_k$ is exactly the space composed of manifold functions that we expect to encode.

\textbf{Encoder.} Assume that \(\{\phi_{n,m}\}_{1\leq n,1\leq m\leq\kappa(n)}\subset C(V)\) is a series of given fixed continuous basis functions defined on $V$ (e.g., orthogonal polynomials, Fourier basis, finite elements). Here $n$ can be regarded as the indexing of the $\kappa(n)$-dimensional subspace ${\rm span}\{\phi_{n,1},...,\phi_{n,\kappa(n)}\}$, and $\kappa(n)\to\infty$ as $n\to\infty$. The encoder is
defined as follows. Given any manifold function
\begin{equation}
    (M,f_M)\in X_k,
\end{equation}
let
\begin{equation}\label{eq:encoder_manifold}
    \Phi_n^1:\mathcal{M}_k\to\mathbb{R}^{\kappa(n)},\quad
    \Phi_n^1(M):=
    \left(\int_M1\cdot\phi_{n,m}\,d\mathcal{H}^k\right)_{1\leq m\leq\kappa(n)}\in\R^{\kappa(n)},
\end{equation}
and
\begin{equation}\label{eq:encoder_function}
    \Phi_n^2:X_k\to\mathbb{R}^{\kappa(n)},\quad
    \Phi_n^2(M,f_M):=
    \left(\int_Mf_M\cdot\phi_{n,m}\,d\mathcal{H}^k\right)_{1\leq m\leq\kappa(n)}\in\R^{\kappa(n)}.
\end{equation}
The final encoder is
\begin{equation}
    \boxed{\Phi_n:X_k\to\R^{2\kappa(n)},\quad \Phi_n(M,f_M):=(\Phi_n^1(M),\Phi_n^2(M,f_M))\in\R^{2\kappa(n)}.}
\end{equation}
Here $\Phi_n^1(M)$ encodes the manifold $M$, and $\Phi_n^2(M,f_M)$ encodes the function $f_M$ defined on $M$. The encoding dimension grows as $n$ increases.

\textbf{Decoder.} For ease of understanding, here we decode the latent codes onto $C(V)$, which simplifies the original setting in \cite{hu2025manifold}. Given a vector $(\gamma_m)_{m=1}^{\kappa(n)}\in\R^{\kappa(n)}$, we denote
\begin{equation}
    \label{eq:single_decoder}\Psi_n^1\left((\gamma_m)_{m=1}^{\kappa(n)}\right):=\sum_{m=1}^{\kappa(n)}\gamma_{m}\phi_{n,m},
\end{equation}
then for the latent vector 
\begin{equation}
    \left((\alpha_{m})_{m=1}^{\kappa(n)},(\beta_{m})_{m=1}^{\kappa(n)}\right)=\Phi_n(M,f_M)\in\R^{2\kappa(n)},
\end{equation}
the decoder can be written as
\begin{equation}
    \boxed{\Psi_n\left((\alpha_{m})_{m=1}^{\kappa(n)},(\beta_{m})_{m=1}^{\kappa(n)}\right):=\left(\Psi_n^1\left((\alpha_m)_{m=1}^{\kappa(n)}\right), \Psi_n^1\left((\beta_m)_{m=1}^{\kappa(n)}\right)\right)\in C(V)\times C(V).}
\end{equation}
Since in the PDE problems we consider, the manifold on which the solution resides is known (for example, it may coincide with one of the input manifolds), it is sufficient to obtain the decoded solution function on the manifold by the restriction operator.

\textbf{Super-algebraic convergence.} The MFE has been proven to achieve super-algebraic convergence based on the commonly used spectral basis functions, such as the Legendre polynomials and the Fourier basis functions.
\begin{theorem}[Corollary 1 and Remark 2 from \cite{hu2025manifold}]
    For Legendre-based (Fourier-based) MFE, the composite of encoder and decoder $\Psi_n\circ\Phi_n$ converges to the identity mapping super-algebraically.
\end{theorem}
The above theorem indicates that MFE can compress the manifold function information into a very low-dimensional vector, which is one of the key reasons why MENO is memory-efficient.

\textbf{MFE tokens.} Given several manifold functions which may be of different dimensions $k_1,...,k_p$,
\begin{equation}\label{eq:input_manifold_functions}
    \begin{split}
        &M_1,f_{M_1}^1,...,f_{M_1}^{l_1-1},\quad (M_1\in\mathcal{M}_{k_1}) \\
        &M_2,f_{M_2}^1,...,f_{M_2}^{l_2-1},\quad (M_2\in\mathcal{M}_{k_2}) \\
        &\qquad\quad\vdots \\
        &M_p,f_{M_p}^1,...,f_{M_p}^{l_p-1}.\quad (M_p\in\mathcal{M}_{k_p})
    \end{split}
\end{equation}
Such manifold functions as the input of PDE problems cover all the possible scenarios. The inputs can be encoded to MFE tokens as
\begin{equation}\label{eq:mfe_tokens}
    \boxed{\left[\Phi_n^{1}(M_1),\Phi_n^2(f_{M_1}^1),...,\Phi_n^2(f_{M_1}^{l_1-1}),...,\Phi_n^1(M_p),\Phi_n^2(f_{M_p}^1),...,\Phi_n^2(f_{M_p}^{l_p-1})\right]\in\R^{\kappa(n)\times(l_1+\cdots+ l_p)},}
\end{equation}
which means a series of $\kappa(n)$ tokens and each token is of dimension $l_1+\cdots+l_p$. The notations $\Phi_n^1$ and $\Phi_n^2$ in Eq. \eqref{eq:mfe_tokens} in fact depend on the dimension $k_i$. To keep the presentation concise, we omit the dependency of notation here.

To help the reader understand, here we show a very concrete example of MFE tokens. Consider the 2-d Poisson equation
\begin{equation}
    \begin{cases}
        -\nabla\cdot(k\nabla u)=f\quad &{\rm in\ }\Omega,\\
        u=g\quad &{\rm on\ }\partial\Omega.
    \end{cases}
\end{equation}
Assume that $\Omega$ is contained in $[0,1]^2$. All the inputs of this problem can be written as
\begin{equation}
    (\Omega, k_\Omega, f_\Omega,\partial\Omega,g_{\partial\Omega}),
\end{equation}
and the Legendre-based MFE tokens are
\begin{equation}\label{eq:example_tokens}
\begin{split}
    \bigg(&\int_{\Omega}1\cdot\ell_{i_1}(x)\cdot\ell_{i_2}(y)dxdy, \int_{\Omega}k_\Omega(x,y)\cdot\ell_{i_1}(x)\cdot\ell_{i_2}(y)dxdy, \int_{\Omega}f_\Omega(x,y)\cdot\ell_{i_1}(x)\cdot\ell_{i_2}(y)dxdy, \\
    &\int_{\partial\Omega}1\cdot\ell_{i_1}(x)\cdot\ell_{i_2}(y)ds,\int_{\partial\Omega}g_{\partial\Omega}(x,y)\cdot\ell_{i_1}(x)\cdot\ell_{i_2}(y)ds\bigg)\in\R^{5},\quad 0\leq i_1,i_2\leq n-1,
\end{split}
\end{equation}
where $\ell_i$ is the $L^2$-normalized Legendre polynomial of degree $i$ on $[0,1]$. Eq. \eqref{eq:example_tokens} shows a series of $n^2$ tokens and each token is of dimension 5. In the architecture of MENO, the dimension of the tokens, i.e. 5, will be lifted to a larger dimension, while the length of the token sequence $n^2$ remains unchanged. The MFE tokens play a key role in the construction of MENO.

\section{Architecture of MENO}
\label{sec:architecture}
The architecture of MENO is composed of two branches: global branch and local branch, as well as an information transfer mechanism between the two branches. Here we consider the Legendre-based MFE as the example, and use the previous $n$ modes of Legendre polynomials for encoding.

\textbf{Global branch.} The global branch is mainly composed of the Transformer blocks. Assume that the input manifold functions of Eq. \eqref{eq:input_manifold_functions} have been encoded to $n^d$ MFE tokens
\begin{equation}\label{eq:input_tokens}
    \underline{Z_{\rm in}\in\R^{n^d\times(l_1+\cdots l_p)},}
\end{equation}
we first apply a lifting layer to $Z_{\rm in}$ together with a bias as
\begin{equation}
    \underline{H_0={\rm Lift(Z_{\rm in})}+\mathcal{B}(P_t)\in\R^{n^d\times h},}
\end{equation}
where $P_t$ is the spectral position information with $t$ frequency modes
\begin{equation}
\begin{split}
    P_t:=&\left(i_1\eta,...,i_d\eta,(\cos(2^s\pi i_1\eta),\sin(2^s\pi i_1\eta),...,\cos(2^s\pi i_d\eta),\sin(2^s\pi i_d\eta))_{s=0}^{t-1}\right)_{0\leq i_1,...,i_d\leq n-1} \\
    &\in\R^{n^d\times(2t+1)d},\quad\eta:=1/(n-1),
\end{split}
\end{equation}
and ${\rm Lift}:\R^{l_1+\cdots l_p}\to\R^h$ is a linear layer, $\mathcal{B}:\R^{(2t+1)d}\to\R^h$ is an MLP (Multilayer Perceptron). Subsequently, we apply the classical Transformer block
\begin{align}
    &\underline{\tilde{H}_{l}
    =H_{l-1}
    +{\rm MHA}_{l}
    \left({\rm LN}(H_{l-1})\right)\in\R^{n^d\times h},} \\
    &\underline{H_{l}
    =\tilde{H}_{l}
    +{\rm MLP}_{l}
    \left({\rm LN}(\tilde{H}_{l})\right)\in\R^{n^d\times h},\quad l=1,...,L,}\label{eq:H_l}
\end{align}
where LN is the layer normalization and MHA is the multi-head attention. After that, we project the dimension of MFE token to the number of output solutions, denoted by $l_{\rm out}$, as
\begin{equation}
    \underline{Z_{\rm out}=n^{-\frac{d}{2}}\cdot{\rm Proj}({\rm LN}(H_L))\in\R^{n^d\times l_{\rm out}},}
\end{equation}
where ${\rm Proj}:\R^h\to\R^{l_{\rm out}}$ is a linear layer, $n^{-\frac{d}{2}}$ is a scaling number for normalizing the variance of Legendre. Lastly, we decode the latent vectors $Z_{\rm out}$ to $l_{\rm out}$ preliminary solutions $\tilde{u}_1^{\rm global},...,\tilde{u}_{l_{\rm out}}^{\rm global}$ by MFE decoder~\eqref{eq:single_decoder}, 
\begin{equation}
    \underline{(\tilde{u}_1^{\rm global},...,\tilde{u}_{l_{\rm out}}^{\rm global})=\left(\Psi_n^1([Z_{\rm out}]_1),...,\Psi_n^1([Z_{\rm out}]_{l_{\rm out}})\right)\in\left(C(V)\right)^{l_{\rm out}},}
\end{equation}
where $[Z_{\rm out}]_i$ means the $i$-th column of $Z_{\rm out}$.

\textbf{Injection.} The subsequent layers are critical for the coming local branch, they receive and process the information from the global branch, and then inject the processed information into the local branch. Denote
\begin{equation}
    \hat{H}_l=n^{-\frac{d}{2}}\cdot{\rm LL}_l^{\rm trans}({\rm LN}(H_l))\in\R^{n^d\times \hat{h}},
\end{equation}
where ${\rm LL}_l^{\rm trans}:\R^{h}\to\R^{\hat{h}}$ are linear layers for $l=1,...,L$, and $H_l\in\R^{n^d\times h}$ is from~\eqref{eq:H_l} in the global branch. Let
\begin{align}
    &\tilde{T}_l(x)=(\Psi_n^1([\hat{H}_l]_1)(x),...,\Psi_n^1([\hat{H}_l]_{\hat{h}})(x))\in\R^{\hat{h}}, \\
    &\tilde{T}_{l}^{i}(x)
    =\left(\frac{\partial}{\partial x_i}\left(\Psi_n^1([\hat{H}_l]_1)\right)(x),...,\frac{\partial}{\partial x_i}\left(\Psi_n^1([\hat{H}_l]_{\hat{h}})\right)(x)\right)\in\R^{\hat{h}},\quad i=1,...,d,
\end{align}
with $x=(x_1,...,x_d)\in V$. $\tilde{T}_l(x)$ and $\tilde{T}_l^i(x)$ are exactly the decoded $\hat{H}_l$ and its derivatives evaluated at $x$. The global-to-local transfer information can be written as
\begin{equation}\label{eq:injection}
    T_l(x)=\tilde{T}_l(x)+W^{\rm grad}_1\cdot\tilde{T}_l^1(x)+\cdots+W^{\rm grad}_d\cdot\tilde{T}_l^d(x)\in\R^{\hat{h}},
\end{equation}
where $W^{\rm grad}_i\in\R^{\hat{h}\times \hat{h}}$ are trainable matrices and all the $L$ layers share the parameters. For brevity, here we only show the case where first-order derivative information is used; in fact, derivative information of arbitrary order can be incorporated if needed.

\textbf{Local branch.} Assume that the solutions $u_1,...,u_{l_{\rm out}}$ are defined on the output manifold $M_0$, i.e. $u_i\in L^2(M_0)$. Note that $M_0$ is not necessarily an element of $\{M_1,...,M_p\}$. Let
\begin{equation}
    \underline{x\in M_0}
\end{equation}
be a query point. We first need to evaluate the input information at $x$ based on~\eqref{eq:input_manifold_functions},~\eqref{eq:input_tokens} and the decoder. Denote the cross evaluation by
\begin{equation}\label{eq:cross_evaluation}
    \tilde{\mathcal{V}}(x):=\left(\Psi_n^1([Z_{\rm in}]_1)(x),...,\Psi_n^1([Z_{\rm in}]_{l_1+\cdots+l_p})(x)\right)\in\R^{l_1+\cdots+l_p}.
\end{equation}
The above $\tilde{\mathcal{V}}(x)$ is built upon the premise that $M_0\notin\{M_1,...,M_p\}$. If $M_0=M_j$ for a $1\leq j\leq p$, then the corresponding values in~\eqref{eq:cross_evaluation} can be replaced by the direct evaluation
\begin{equation}
    (1,f_{M_j}^1(x),...,f_{M_j}^{l_j-1}(x)).
\end{equation}
Therefore we denote the final evaluation as
\begin{equation}
    \underline{\mathcal{V}(x)=\left({\rm Modify}(\tilde{\mathcal{V}}(x)),P_t^{\rm local}(x)\right)\in\R^{l_1+\cdots+l_p+(2t+1)d},}
\end{equation}
by checking the modification and additionally concatenating a position information
\begin{equation}
    P_t^{\rm local}(x):=\left(x,(\cos(2^s\pi x_1),\sin(2^s\pi x_1),...,\cos(2^s\pi x_d),\sin(2^s\pi x_d))_{s=0}^{t-1}\right)\in\R^{(2t+1)d},
\end{equation}
with $x=(x_1,...,x_d)$. Similar to the global branch, we also apply a lifting layer to $\mathcal{V}(x)$ as
\begin{equation}
    \underline{H_0^{\rm local}(x)={\rm Lift}^{\rm local}(\mathcal{V}(x))\in\R^{\hat{h}},}
\end{equation}
and ${\rm Lift}^{\rm local}:\R^{l_1+\cdots+l_p+(2t+1)d}\to\R^{\hat{h}}$ is a linear layer. 

With $T_l(x)$ from~\eqref{eq:injection}, the information from the global branch can be injected into this local branch, and the next layers are shown as
\begin{equation}
    \underline{H_l^{\rm local}(x)=\sigma \left({\rm LL}_l^{\rm local}({\rm LN}(H_{l-1}^{\rm local}(x)))+T_l(x)\right)\in\R^{\hat{h}},\quad l=1,...,L,}
\end{equation}
and ${\rm LL}_l^{\rm local}:\R^{\hat{h}}\to\R^{\hat{h}}$ are linear layers, $\sigma$ is the activation function. The final output is
\begin{equation}
    \underline{(\tilde{u}_1^{\rm local}(x),...,\tilde{u}_{l_{\rm out}}^{\rm local}(x))={\rm Proj}^{\rm local}({\rm LN}(H_L^{\rm local}(x)))\in\R^{l_{\rm out}},}
\end{equation}
where ${\rm Proj}^{\rm local}:\R^{\hat{h}}\to\R^{l_{\rm out}}$ is also a linear layer.

\begin{figure}[htbp]
    \centering
    \resizebox{\textwidth}{!}{%
    \begin{tikzpicture}[
      x=1cm,
      y=1cm,
      font=\sffamily\scriptsize,
      text=ink,
      >={Latex[length=2.2mm,width=1.5mm]},
      flow/.style={->,draw=ink,line width=0.85pt,rounded corners=2.5pt},
      global flow/.style={flow,draw=globalLine},
      local flow/.style={flow,draw=localLine},
      analytic flow/.style={flow,draw=mfeLine},
      injection flow/.style={flow,draw=mfeLine},
      state/.style={
        rectangle,rounded corners=2pt,draw=rule,fill=paper,
        line width=0.75pt,align=center,minimum width=.0cm,
        minimum height=8mm,inner xsep=3pt,inner ysep=3pt
      },
      input/.style={
        state,draw=ink,line width=0.95pt,minimum width=1cm,
        minimum height=1mm,font=\sffamily\footnotesize
      },
      input global/.style={
        state,draw=globalLine,fill=globalFill,line width=0.95pt
      },
      input local/.style={
        state,draw=localLine,fill=localFill,line width=0.95pt
      },
      analytic/.style={
        rectangle,rounded corners=8pt,draw=mfeLine,fill=mfeFill,
        line width=0.95pt,align=center,
        minimum width=1cm,minimum height=8mm,inner xsep=4pt
      },
      linear global/.style={
    	trapezium,trapezium left angle=76,trapezium right angle=104,
    	draw=globalLine,fill=globalFill,line width=0.95pt,
    	align=center,minimum width=1.0cm,minimum height=7mm,inner xsep=0pt
    },
    linear local/.style={
    	trapezium,trapezium left angle=76,trapezium right angle=104,
    	draw=localLine,fill=localFill,line width=0.95pt,
    	align=center,minimum width=1.0cm,minimum height=7mm,inner xsep=0pt
    },
      encoding trapezoid/.style={
        trapezium left angle=82,trapezium right angle=82,
        shape border rotate=90,minimum width=12.7mm,minimum height=18mm,
        inner xsep=-1pt,inner ysep=-3.5pt
      },
      decoding trapezoid/.style={
        trapezium left angle=82,trapezium right angle=82,
        shape border rotate=270,minimum width=12.7mm,minimum height=18mm,
        inner xsep=-1pt,inner ysep=-1pt
      },
      preprocess local/.style={
        rectangle,draw=localLine,fill=localFill,line width=0.9pt,
        align=center,minimum width=2.55cm,minimum height=8mm,inner xsep=5pt
      },
      sum/.style={
        circle,draw=ink,fill=paper,line width=0.65pt,
        minimum size=5mm,inner sep=0pt,font=\sffamily\normalsize
      },
      lane label/.style={font=\sffamily\bfseries\scriptsize,align=center},
      note/.style={font=\sffamily\fontsize{6.5}{7.2}\selectfont,
        text=muted,align=center},
      group/.style={
        rectangle,rounded corners=5pt,draw=rule,fill=panel,
        line width=0.75pt,inner sep=7pt
      },
      section tag/.style={
        font=\sffamily\bfseries\scriptsize,fill=paper,text=ink,
        inner xsep=6pt,inner ysep=2pt
      },
      process global/.style={
        state,draw=globalLine,fill=globalFill,minimum width=4.10cm,
        minimum height=17mm
      },
      process local/.style={
        state,draw=localLine,fill=localFill,minimum width=4.10cm,
        minimum height=17mm
      },
      main global op/.style={
        state,draw=globalLine,fill=globalFill,minimum width=1.15cm,
        minimum height=10mm,inner xsep=3pt
      },
      main local op/.style={
        state,draw=localLine,fill=localFill,minimum width=1.15cm,
        minimum height=10mm,inner xsep=3pt
      },
      main panel/.style={
      	rectangle,rounded corners=5pt,draw=ink,fill=panel,
      	line width=0.7pt,inner sep=8pt
      },
      shadow panel/.style={
      	rectangle,rounded corners=5pt,draw=rule,fill=shadow,
      	line width=0.7pt,inner sep=8pt
      },
      detail ln global/.style={
        rectangle,rounded corners=7pt,draw=globalLine,fill=globalFill,
        line width=0.9pt,align=center,minimum width=2.05cm,
        minimum height=5.5mm,inner xsep=3pt
      },
      detail ln local/.style={
        rectangle,rounded corners=7pt,draw=localLine,fill=localFill,
        line width=0.9pt,align=center,minimum width=2.05cm,
        minimum height=5.5mm,inner xsep=3pt
      },
      detail attention/.style={
        rectangle,rounded corners=1pt,draw=globalLine,fill=globalFill,
        line width=0.9pt,align=center,minimum width=2.55cm,
        minimum height=6mm,inner xsep=3pt
      },
      detail mlp/.style={
        rectangle,rounded corners=2pt,draw=globalLine,fill=globalFill,
        line width=0.9pt,align=center,minimum width=2.15cm,
        minimum height=6mm,inner xsep=3pt
      },
      detail activation/.style={
        ellipse,draw=localLine,fill=localFill,line width=0.9pt,
        minimum width=1.0cm,minimum height=5mm,inner sep=0.5pt
      },
      detail mixer/.style={
        trapezium,trapezium left angle=78,trapezium right angle=102,
        draw=gradLine,fill=gradFill,line width=0.9pt,align=center,
        minimum width=1.7cm,minimum height=7mm,inner xsep=5pt
      },
      repeat badge/.style={
        rectangle,rounded corners=3pt,fill=ink,text=paper,
        inner xsep=8pt,inner ysep=3pt,font=\sffamily\bfseries\scriptsize
      }
    ]
    
    \node[input,minimum width=3cm,minimum height=5mm] (manifoldInput) at (0,-.6) {
      $M_1,f_{M_1}^1,...,f_{M_1}^{l_1-1}, $\\[2pt]
      $\vdots$\\[-1pt]
      $M_p,f_{M_p}^1,...,f_{M_p}^{l_p-1}.$
    };
    
    \node[analytic] (mfeEncoder) at (2.70,.95)
      {MFE Encoder};
    
    \node[state, draw=globalLine,minimum width=1cm,minimum height=17mm]
      (globalMoments) at (5.85,.95) {
      $\begin{bmatrix}
      	\Phi_n^{1}(M_1)\\[-1pt]
      	\Phi_n^2(f_{M_1}^1)\\[-1pt]
      	\vdots\\[-1pt]
      	\Phi_n^1(M_p)\\[-1pt]
      	\vdots\\[-1pt]
      	\Phi_n^2(f_{M_p}^{l_p-1})
      \end{bmatrix}$
    };
    
    \node[linear global,encoding trapezoid] (globalLifting) at (11.5,.95)
      {Global\\Lifting};
    
    \node[input global,minimum width=3.15cm] (positionBias) at (9.,-.2) {
      Additive Position Bias
    };
    
    \node[state,draw=globalLine,line width=1.0pt,minimum width=2.45cm]
      (globalState) at (14.55,.95) {
      Global State
    };
    
    \coordinate (globalAend) at (16.5,.95);
    \coordinate (globalMix) at (12.85,.95);
    
    \draw[global flow] (manifoldInput.north) |- (mfeEncoder.west);
    \draw[global flow] (mfeEncoder) -- (globalMoments);
    \draw[global flow] (globalMoments) -- (globalLifting);
    \draw[global flow] (positionBias) -| (globalMix.south);
    \draw[global flow] (globalLifting) -- (globalState);
    \draw[global flow, dashed] (globalState) -- (globalAend);
    
    \node[input,minimum width=3cm,minimum height=5mm] (queryCoordinates) at (0,-3.3) {
    	Query $x\in M_0$
    };
    
    \node[preprocess local,minimum width=2.35cm] (evaluate) at (2.60,-2.2) {
      Direct Evaluation
    };
    
    \node[analytic] (mfeDecoder) at (5.85,-1.155) {
      Cross Evaluation\\(MFE Decoder)
    };
    
    \node[state,draw=localLine,minimum width=1.0cm,minimum height=17mm]
      (pointFeatures) at (5.85,-2.75) {
      $\begin{bmatrix}
      	\mathcal{V}(x)
      \end{bmatrix}$
    };
    
    \node[input local,minimum width=3.15cm] (fourierFeatures) at (9.,-1.6) {
      Position Concatenation
    };
    
    \node[linear local,encoding trapezoid] (localLifting) at (11.5,-2.75) {
      Local\\Lifting
    };
    
    \node[state,draw=localLine,line width=1.0pt,minimum width=2.45cm]
      (localState) at (14.55,-2.75) {
      Local State
    };
    
    \coordinate (localAend) at (16.5,-2.75);

    \draw[local flow, dashed] (manifoldInput.south) |- (evaluate.west);
    \draw[local flow, dashed] (evaluate.east) -- ($(pointFeatures.west)+(0,0.55)$);
    \draw[local flow] (queryCoordinates.east) -- ($(pointFeatures.west)-(0,0.55)$);
    \draw[analytic flow] (globalMoments.south) -- (mfeDecoder.north);
    \draw[analytic flow] (mfeDecoder) -- (pointFeatures.north);
    \draw[local flow] (pointFeatures) -- (localLifting);
    \draw[local flow] (fourierFeatures.south) |- ($(pointFeatures.east)+(0,0.4)$);
    \draw[local flow] (localLifting) -- (localState);
    \draw[local flow, dashed] (localState) -- (localAend);

    \draw[draw=ink,line width=0.95pt] (-1.75,-4.45) -- (16.95,-4.45);
    \node[section tag,anchor=south west,font=\sffamily\bfseries] at (-1.65,-4.40)
      {1 \quad Encoding};
    \node[section tag,anchor=north west,font=\sffamily\bfseries] at (-1.65,-4.52)
      {2 \quad Processing and Decoding};
    
    \node[state,draw=globalLine,line width=1.0pt,minimum width=1.95cm,inner xsep=6pt]
      (procGlobalInput) at (0.2,-6.75) {
      Global State
    };
    \node[state,draw=localLine,line width=1.0pt,minimum width=1.95cm,inner xsep=6pt]
      (procLocalInput) at (0.2,-10.45) {
      Local State
    };
    
    \node[main global op,minimum width=1.05cm] (attentionOne) at (2.80,-6.75) {Attention};
    \node[main global op,minimum width=1.05cm] (mlpOne) at (5.2,-6.75) {MLP};
    \node[main local op,minimum width=1.05cm] (localLinearOne) at (2.60,-10.45) {Linear};
    \node[sum,draw=localLine] (localSumOne) at (3.9,-10.45) {$+$};
    \node[analytic,minimum width=1.75cm] (reconstructionOne) at (3.9,-8.6) {
      Injection
    };
    \node[detail activation] (sigma) at (5.2,-10.45) {$\sigma$};
    
    \coordinate (glWest) at (1.8,-6.2);
    \coordinate (glPanelEast) at (5.9,-11.05);
    
    \begin{scope}[on background layer]
    	\node[shadow panel,fit=(attentionOne)(mlpOne)(localLinearOne)(localSumOne)(reconstructionOne)(sigma)(glWest)(glPanelEast),
    	xshift=3mm,yshift=2mm] (layerShadowTwo) {};
    	\node[shadow panel,fit=(attentionOne)(mlpOne)(localLinearOne)(localSumOne)(reconstructionOne)(sigma)(glWest)(glPanelEast),
    	xshift=1.5mm,yshift=1mm] (layerShadowOne) {};
    	\node[main panel,fit=(attentionOne)(mlpOne)(localLinearOne)(localSumOne)(reconstructionOne)(sigma)(glWest)(glPanelEast)] (Pair) {};
    \end{scope}
    
    \node[lane label,text=ink,anchor=south] at
      ($(Pair.north)+(1.5mm,2mm)$) {Global-to-Local Block $\times L$};
      
    \coordinate (globalBstart) at (-1.3,-6.75);
    \coordinate (localBstart) at (-1.3,-10.45);
    
    \draw[global flow, dashed] (globalBstart) -- (procGlobalInput);
    \draw[global flow, dashed] (procGlobalInput) -- (attentionOne);
    \draw[global flow] (attentionOne) -- (mlpOne);

    \draw[local flow, dashed] (localBstart) -- (procLocalInput);
    \draw[local flow, dashed] (procLocalInput) -- (localLinearOne);
    \draw[local flow] (localLinearOne) --  (localSumOne);
    \draw[local flow] (localSumOne) --  (sigma);

    \draw[injection flow] (mlpOne.south) |- (reconstructionOne.east);
    \draw[injection flow] (reconstructionOne.south) -- ++(0,-0.45) -| (localSumOne.north);
    
    \node[linear global,decoding trapezoid]
      (globalReadout) at (8.,-6.75) {
      Global\\Projection
    };
    \node[analytic]
      (globalOutputDecode) at (10.6,-6.75) {
      MFE Decoder
    };
    \node[state,draw=globalLine,minimum width=2.25cm,inner xsep=6pt]
      (globalPrediction) at (13.5,-6.75) {
      Global Prediction
    };
    
    \node[linear local,decoding trapezoid]
      (localReadout) at (8.,-10.45) {
      Local\\Projection
    };
    \node[state,draw=localLine,minimum width=2.25cm,inner xsep=6pt]
      (localPrediction) at (11.20,-10.45) {
      Local Prediction
    };
    
    \node[sum] (finalAdd) at (13.5,-8.8) {$+$};
    \node[state,draw=ink,line width=1.05pt,minimum width=1.80cm,
      minimum height=20mm,inner xsep=3pt]
      (finalPrediction) at (15.60,-8.8) {
      $\begin{bmatrix}
      	\tilde{u}_1(x)\\\vdots\\\tilde{u}_{l_{\rm out}}(x)
      \end{bmatrix}$
    };
    
    \draw[global flow,densely dashed] (mlpOne) -- (globalReadout);
    \draw[global flow] (globalReadout) -- (globalOutputDecode);
    \draw[global flow] (globalOutputDecode) -- (globalPrediction);

    \draw[local flow,densely dashed] (sigma) -- (localReadout);
    \draw[local flow] (localReadout) -- (localPrediction);
    \draw[global flow] (globalPrediction.south) -- (finalAdd.north);
    \draw[local flow] (localPrediction.east) -| (finalAdd.south);
    \draw[flow] (finalAdd) -- (finalPrediction);
    
    \end{tikzpicture}
    }
    \caption{\textbf{MENO architecture.} The overall architecture is composed of a global branch and a local branch. In the repeated global-to-local blocks, the information of the global branch is injected into the local branch. The two branch outputs are summed to yield the final result.}
    \label{fig:meno_architecture}
\end{figure}
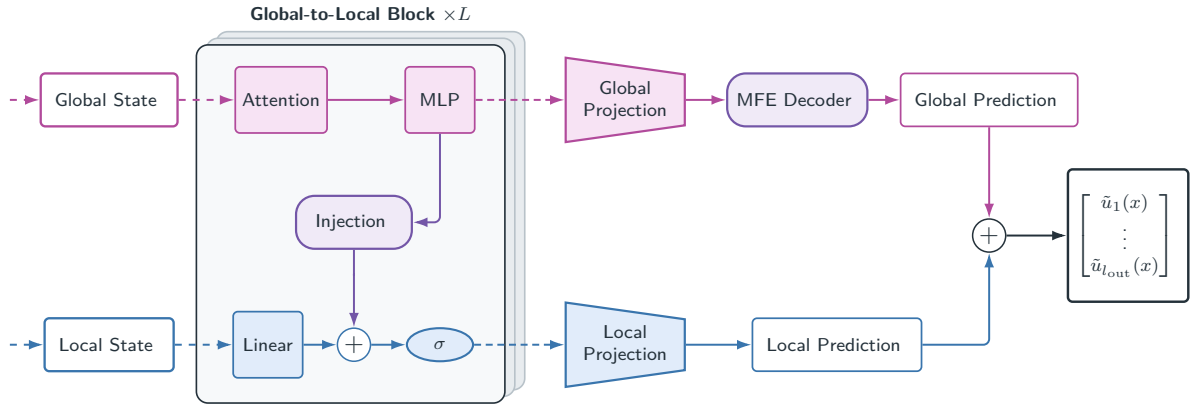

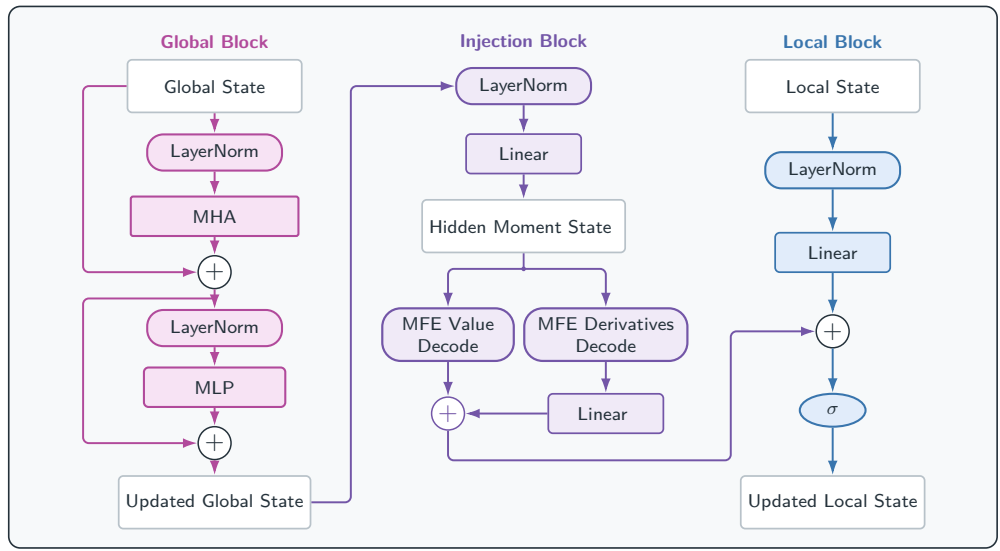
\begin{figure}[htbp]
    \centering
    \resizebox{.8\textwidth}{!}{%
    \begin{tikzpicture}[
      x=1cm,
      y=1cm,
      font=\sffamily\scriptsize,
      text=ink,
      >={Latex[length=2.2mm,width=1.5mm]},
      flow/.style={->,draw=ink,line width=0.85pt,rounded corners=2.5pt},
      global flow/.style={flow,draw=globalLine},
      local flow/.style={flow,draw=localLine},
      analytic flow/.style={flow,draw=mfeLine},
      injection flow/.style={flow,draw=mfeLine},
      state/.style={
        rectangle,rounded corners=2pt,draw=rule,fill=paper,
        line width=0.75pt,align=center,minimum width=.0cm,
        minimum height=8mm,inner xsep=3pt,inner ysep=3pt
      },
      input/.style={
        state,draw=ink,line width=0.95pt,minimum width=1cm,
        minimum height=1mm,font=\sffamily\footnotesize
      },
      input global/.style={
        state,draw=globalLine,fill=globalFill,line width=0.95pt
      },
      input local/.style={
        state,draw=localLine,fill=localFill,line width=0.95pt
      },
      analytic/.style={
        rectangle,rounded corners=8pt,draw=mfeLine,fill=mfeFill,
        line width=0.95pt,align=center,
        minimum width=1cm,minimum height=8mm,inner xsep=4pt
      },
      linear global/.style={
    	trapezium,trapezium left angle=76,trapezium right angle=104,
    	draw=globalLine,fill=globalFill,line width=0.95pt,
    	align=center,minimum width=1.0cm,minimum height=7mm,inner xsep=-3pt
    },
    linear local/.style={
    	trapezium,trapezium left angle=76,trapezium right angle=104,
    	draw=localLine,fill=localFill,line width=0.95pt,
    	align=center,minimum width=1.0cm,minimum height=7mm,inner xsep=-3pt
    },
      encoding trapezoid/.style={
        trapezium left angle=82,trapezium right angle=82,
        shape border rotate=90,minimum width=12.7mm,minimum height=18mm,
        inner xsep=-1pt,inner ysep=-3.5pt
      },
      decoding trapezoid/.style={
        trapezium left angle=82,trapezium right angle=82,
        shape border rotate=270,minimum width=12.7mm,minimum height=18mm,
        inner xsep=-1pt,inner ysep=-1pt
      },
      preprocess local/.style={
        rectangle,draw=localLine,fill=localFill,line width=0.9pt,
        align=center,minimum width=2.55cm,minimum height=8mm,inner xsep=5pt
      },
      sum/.style={
        circle,draw=ink,fill=paper,line width=0.65pt,
        minimum size=5mm,inner sep=0pt,font=\sffamily\normalsize
      },
      lane label/.style={font=\sffamily\bfseries\scriptsize,align=center},
      note/.style={font=\sffamily\fontsize{6.5}{7.2}\selectfont,
        text=muted,align=center},
      group/.style={
        rectangle,rounded corners=5pt,draw=rule,fill=panel,
        line width=0.75pt,inner sep=7pt
      },
      section tag/.style={
        font=\sffamily\bfseries\scriptsize,fill=paper,text=ink,
        inner xsep=6pt,inner ysep=2pt
      },
      process global/.style={
        state,draw=globalLine,fill=globalFill,minimum width=4.10cm,
        minimum height=17mm
      },
      process local/.style={
        state,draw=localLine,fill=localFill,minimum width=4.10cm,
        minimum height=17mm
      },
      main global op/.style={
        state,draw=globalLine,fill=globalFill,minimum width=1.15cm,
        minimum height=10mm,inner xsep=3pt
      },
      main local op/.style={
        state,draw=localLine,fill=localFill,minimum width=1.15cm,
        minimum height=10mm,inner xsep=3pt
      },
      main panel/.style={
      	rectangle,rounded corners=5pt,draw=ink,fill=panel,
      	line width=0.7pt,inner sep=8pt
      },
      shadow panel/.style={
      	rectangle,rounded corners=5pt,draw=rule,fill=shadow,
      	line width=0.7pt,inner sep=8pt
      },
      detail ln global/.style={
        rectangle,rounded corners=7pt,draw=globalLine,fill=globalFill,
        line width=0.9pt,align=center,minimum width=2.05cm,
        minimum height=5.5mm,inner xsep=3pt
      },
      detail ln local/.style={
        rectangle,rounded corners=7pt,draw=localLine,fill=localFill,
        line width=0.9pt,align=center,minimum width=2.05cm,
        minimum height=5.5mm,inner xsep=3pt
      },
      detail attention/.style={
        rectangle,rounded corners=1pt,draw=globalLine,fill=globalFill,
        line width=0.9pt,align=center,minimum width=2.55cm,
        minimum height=6mm,inner xsep=3pt
      },
      detail mlp/.style={
        rectangle,rounded corners=2pt,draw=globalLine,fill=globalFill,
        line width=0.9pt,align=center,minimum width=2.15cm,
        minimum height=6mm,inner xsep=3pt
      },
      detail activation/.style={
        ellipse,draw=localLine,fill=localFill,line width=0.9pt,
        minimum width=1.0cm,minimum height=5mm,inner sep=0.5pt
      },
      detail mixer/.style={
        trapezium,trapezium left angle=78,trapezium right angle=102,
        draw=gradLine,fill=gradFill,line width=0.9pt,align=center,
        minimum width=1.7cm,minimum height=7mm,inner xsep=5pt
      },
      repeat badge/.style={
        rectangle,rounded corners=3pt,fill=ink,text=paper,
        inner xsep=8pt,inner ysep=3pt,font=\sffamily\bfseries\scriptsize
      }
    ]
    
    \node[lane label,text=globalLine] (detailGlobalLabel) at (3.00,-12.)
      {Global Block};
    \node[lane label,text=mfeLine] (detailInjectionLabel) at (7.70,-12.)
      {Injection Block};
    \node[lane label,text=localLine] (detailLocalLabel) at (12.40,-12.)
      {Local Block};
    
    \node[state,minimum width=2.65cm] (detailGlobalInput) at (3.00,-12.67)
      {Global State};
    \node[detail ln global] (detailGlobalLnOne) at (3.00,-13.68)
      {LayerNorm};
    \node[detail attention] (detailGlobalAttention) at (3.00,-14.65)
      {MHA};
    \node[sum] (detailGlobalAddOne) at (3.00,-15.50) {$+$};
    \node[detail ln global] (detailGlobalLnTwo) at (3.00,-16.35)
      {LayerNorm};
    \node[detail mlp] (detailGlobalMlp) at (3.00,-17.25)
      {MLP};
    \node[sum] (detailGlobalAddTwo) at (3.00,-18.10) {$+$};
    \node[state,minimum width=2.65cm] (detailGlobalOutput) at (3.00,-19.00)
      {Updated Global State};
    
    \draw[global flow] (detailGlobalInput) -- (detailGlobalLnOne);
    \draw[global flow] (detailGlobalLnOne) -- (detailGlobalAttention);
    \draw[global flow] (detailGlobalAttention) -- (detailGlobalAddOne);
    \draw[global flow] (detailGlobalAddOne) -- (detailGlobalLnTwo);
    \draw[global flow] (detailGlobalLnTwo) -- (detailGlobalMlp);
    \draw[global flow] (detailGlobalMlp) -- (detailGlobalAddTwo);
    \draw[global flow] (detailGlobalAddTwo) -- (detailGlobalOutput);
    
    \coordinate (detailSecondResidualTap) at (3.00,-15.90);
    \coordinate (detailResidualExtent) at (0.15,-15.90);
    \draw[global flow] (detailGlobalInput.west) -- (1.00,-12.67)
      |- (detailGlobalAddOne.west);
    \fill[globalLine] (detailSecondResidualTap) circle[radius=1.0pt];
    \draw[global flow] (detailSecondResidualTap) -- (1.00,-15.90)
      |- (detailGlobalAddTwo.west);
    
    \node[detail ln global,draw=mfeLine,fill=mfeFill]
      (detailInjectionLn) at (7.70,-12.67) {LayerNorm};
    \node[main global op,draw=mfeLine,fill=mfeFill,minimum width=1.75cm,
      minimum height=6mm]
      (detailInjectionLinear) at (7.70,-13.70) {Linear};
    \node[state,minimum width=2.75cm]
      (detailHiddenMoments) at (7.70,-14.80) {
      Hidden Moment State
    };
    
    \draw[injection flow] (detailGlobalOutput.east) -- ++(0.60,0)
      |- (detailInjectionLn.west);
    \draw[injection flow] (detailInjectionLn) -- (detailInjectionLinear);
    \draw[injection flow] (detailInjectionLinear) -- (detailHiddenMoments);
    
    \coordinate (detailCoefficientSplit) at (7.70,-15.45);
    \node[analytic,minimum width=2.00cm,text width=1.50cm]
      (detailValueDecode) at (6.55,-16.45) {
      MFE Value\\Decode
    };
    \node[analytic,minimum width=2.30cm,text width=2.20cm]
      (detailGradientDecode) at (8.95,-16.45) {
      MFE Derivatives\\Decode
    };
    \node[main global op,draw=mfeLine,fill=mfeFill,minimum width=1.75cm,
    minimum height=6mm] (detailGradientLinear) at (8.95,-17.65) {
      Linear
    };
    \node[sum,draw=mfeLine,text=mfeLine]
      (detailInjectionAdd) at (6.55,-17.65) {$+$};
    
    \draw[draw=mfeLine,line width=0.8pt]
      (detailHiddenMoments.south) -- (detailCoefficientSplit);
    \fill[mfeLine] (detailCoefficientSplit) circle[radius=1.0pt];
    \draw[injection flow] (detailCoefficientSplit) -| (detailValueDecode.north);
    \draw[injection flow] (detailCoefficientSplit) -| (detailGradientDecode.north);
    \draw[injection flow] (detailValueDecode) -- (detailInjectionAdd);
    \draw[injection flow] (detailGradientDecode) -- (detailGradientLinear);
    \draw[injection flow] (detailGradientLinear.west) -- (detailInjectionAdd.east);
    
    \node[state,minimum width=2.65cm] (detailLocalInput) at (12.40,-12.67)
      {Local State};
    \node[detail ln local] (detailLocalLn) at (12.40,-13.95)
      {LayerNorm};
    \node[main local op,minimum width=1.75cm,minimum height=6mm]
      (detailLocalLinear) at (12.40,-15.2) {Linear};
    \node[sum] (detailLocalAdd) at (12.40,-16.4) {$+$};
    \node[detail activation] (detailLocalSigma) at (12.40,-17.60)
      {$\sigma$};
    \node[state,minimum width=2.65cm] (detailLocalOutput) at (12.40,-19.00)
      {Updated Local State};
    
    \draw[local flow] (detailLocalInput) -- (detailLocalLn);
    \draw[local flow] (detailLocalLn) -- (detailLocalLinear);
    \draw[local flow] (detailLocalLinear) -- (detailLocalAdd);
    \draw[injection flow] (detailInjectionAdd.south) -- ++(0,-0.45)
      -- ++(4.30,0) |- (detailLocalAdd.west);
    \draw[local flow] (detailLocalAdd) -- (detailLocalSigma);
    \draw[local flow] (detailLocalSigma) -- (detailLocalOutput);
    
    \coordinate (detailPanelWest) at (1.02,-12.05);
    \coordinate (detailPanelEast) at (14.62,-19.35);
    \begin{scope}[on background layer]
      \node[main panel,
        fit=(detailGlobalLabel)(detailInjectionLabel)(detailLocalLabel)
            (detailGlobalInput)(detailGlobalOutput)(detailLocalInput)
            (detailLocalOutput)(detailGradientLinear)(detailResidualExtent)
            (detailPanelWest)(detailPanelEast)] {};
    \end{scope}
    
    \end{tikzpicture}
    }
    \caption{\textbf{Global-to-local block.} A unidirectional information injection mechanism.}
    \label{fig:g2l_block}
\end{figure}

\textbf{Merge and output.} Lastly, we merge the global and local branches and output the result solutions as
\begin{equation}
    \underline{(\tilde{u}_1(x),...,\tilde{u}_{l_{\rm out}}(x))=(\tilde{u}_1^{\rm global}(x),...,\tilde{u}_{l_{\rm out}}^{\rm global}(x))+(\tilde{u}_1^{\rm local}(x),...,\tilde{u}_{l_{\rm out}}^{\rm local}(x))\in\R^{l_{\rm out}}.}
\end{equation}
We provide a comprehensive illustration of the proposed MENO architecture in Figure \ref{fig:meno_architecture}. A detailed diagram of the global-to-local block is shown in Figure \ref{fig:g2l_block}.

\textbf{Statistical normalization (optional).} We map the physical domains to \(V=[0,1]^d\) using a common affine transformation, with scales either determined from prior knowledge of the dataset or estimated from the training data. In the latter case, we enlarge the enclosing box by a factor slightly larger than 1 (e.g. \(1.02\)) to allow some margin for test geometries. We may further standardize the inputs using means and standard deviations computed from the training data and kept fixed thereafter. For the global branch, these statistics are computed across training examples separately for each mode and channel of \(Z_{\rm in}\). For the local branch, the function values in \(\mathcal{V}(x)\) are standardized channel-wise before lifting, with points weighted equally within each training example and examples weighted equally across the dataset. Normalization is optional, and selected channels, such as the local unit normal vectors, are left unchanged. Output statistics are computed in the same way as local function statistics for each solution channel. When output normalization is enabled, both branches contribute in the same normalized solution space, and their sum is transformed back to the original units:
\begin{equation}
\tilde{u}_i(x)=\mu_i^{\rm out}
+s_i^{\rm out}\left(
\tilde{u}_i^{\rm global}(x)+\tilde{u}_i^{\rm local}(x)
\right),\qquad i=1,\ldots,l_{\rm out},
\end{equation}
where \(\mu_i^{\rm out}\) and \(s_i^{\rm out}\) are the training mean and standard deviation of \(u_i\). Thus, inverse normalization is applied once after merging the two branches.

\textbf{Why memory-efficient.} The FNO series, the Transolver series, as well as others, are all constructed based on the integration mechanism, such as the kernel function integration or the physical attention integration. Since the integration will involve all the physical space points in the mesh or the point cloud, it causes an extremely large cache usage and computation cost. The overall architecture of MENO is without any integration operator at the training stage, thus the GPU memory usage is independent of the data resolution, and this significantly reduces the memory footprint, especially for large-scale problems.

\textbf{Cross-geometry and general PDE neural solver.} The setting of cross-geometry allows MENO to behave as a general PDE neural solver, which accepts all the possible PDE inputs. Taking the real world elasticity problem as an example, the full inputs may contain: the geometry and material properties (3-d manifold function), the force boundary condition (0/1/2-d manifold function), the displacement boundary condition (0/1/2-d manifold function), the contact and friction condition (2-d manifold function), and so on. All the inputs can be unified and efficiently encoded via the MFE in the MENO. An extension strategy like SDF also makes other neural operators possible to train a general PDE neural solver, however, this will be prohibitively expensive in terms of memory and training cost.

\section{Numerical Experiments}
\label{sec:experiments}
In this section, we compare the MENO with the recent popular neural operators for varying geometry setting, with a primary focus on the Transolver-3 and the PCNO. To rule out the effect of insufficient tuning, we directly compare against the results reported in the original papers of these neural operators. All the experiments are run on an NVIDIA A100 PCIe GPU.

\textbf{Common setting.} In the following experiments, we apply the Legendre-based MFE. The optimizer is chosen as AdamW \cite{loshchilov2017decoupled}, and the activation is chosen as GeLU \cite{hendrycks2016gaussian}. For each case, we train the MENO for 500 epochs and select the model achieving the highest validation accuracy as the final checkpoint. If the dataset is without the validation set, we directly select the model achieving the highest test accuracy as the checkpoint. When comparing peak training GPU memory usage, we adopt the same batch size and the same number of query points, see Table \ref{tab:data_protocol}. Other detailed training settings of MENO for each case can be found in Table \ref{tab:model_config}.

\begin{table}[H]
    \centering
    \footnotesize
    \begin{tabular}{lrrrcl}
        \toprule
        Dataset & Train & Val & Test & Batch size & Train queries  \\
        \midrule
        Poisson (cross-geometry) & 9000 & 0 & 1000 & 10 & 10000 (all)  \\
        Poisson (single-geometry) & 1000 & 200 & 400 & 4 & 2500 dynamic \\
        NASA--CRM & 105 & 0 & 44 & 1 & 16384 dynamic  \\
        AhmedML & 400 & 50 & 50 & 1 & 16384 dynamic \\
        \bottomrule
    \end{tabular}
    \caption{\textbf{Dataset partitions and training protocols.} When comparing peak training GPU memory usage, we adopt the same batch size and the same number of query points.}
    \label{tab:data_protocol}
\end{table}

\begin{table}[H]
    \centering
    \scriptsize
    \begin{tabular}{lccccccc}
        \toprule
        Case & Modes $n$ & Layers $L$ & Hidden dim. $h$/$\hat{h}$ & GF/LF & \#Heads & Width & SN \\
        \midrule
        Poisson (cross-geometry) & 12 & 4 & 128/128 & 4/6 & 8 & 256 & on \\
        Poisson (single-geometry) & 32 & 4 & 160/160 & 8/2 & 10 & 320 & on \\
        NASA-CRM & 8 & 6 & 512/512 & 4/6 & 8 & 512 & on \\
        AhmedML & 16 & 6 & 512/512 & 4/6 & 8 & 256 & on \\
        AhmedML (small) & 8 & 4 & 256/256 & 4/6 & 8 & 256 & on \\
        \bottomrule
    \end{tabular}
    \caption{\textbf{Setting of MENO for each case.} Here ``GF'' and ``LF'' are global and local positional Fourier frequencies, ``\#Heads'' is the number of heads in the MHA, ``Width'' is the width of MLP following MHA, ``SN'' is the statistical normalization.}
    \label{tab:model_config}
\end{table}

\subsection{Poisson equation.} 
\begin{figure}[htbp]
    \centering
    \includegraphics[width=.99\textwidth]{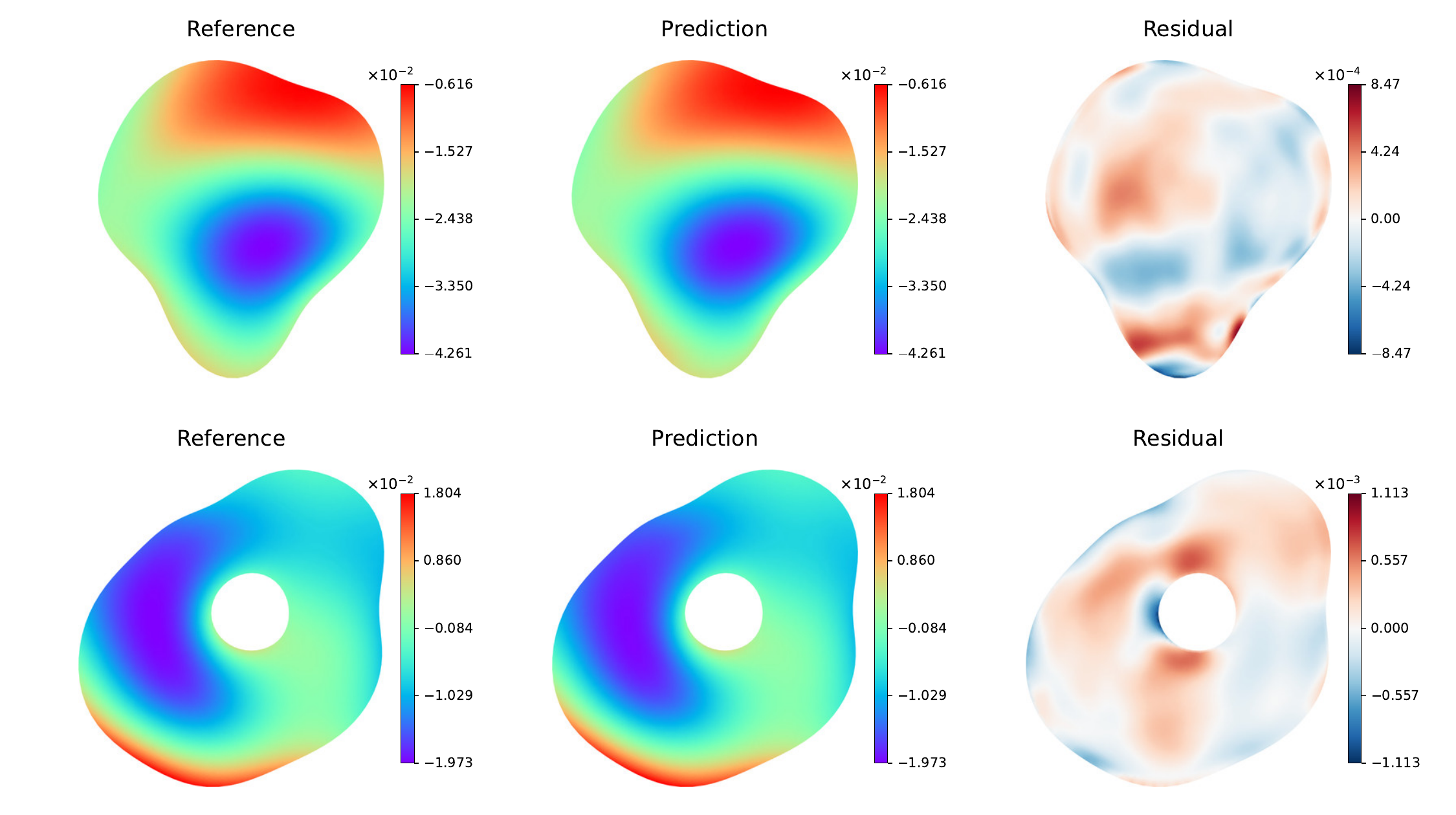}
    \caption{\textbf{Prediction of Poisson equation (cross-geometry).}}
    \label{fig:poisson}
\end{figure}

The first case is the 2-d Poisson equation with variable coefficients
\begin{equation}
    \begin{cases}
        -\nabla\cdot(k\nabla u)=f\quad &{\rm in\ }\Omega,\\
        u=g\quad &{\rm on\ }\partial\Omega,
    \end{cases}
\end{equation}
where the solution mapping is
\begin{equation}
    \mathcal{G}_{\mathrm{Poisson}}^{\rm cross}:\left(\Omega, k_\Omega, f_\Omega,\partial\Omega,g_{\partial\Omega}\right)\mapsto u_\Omega.
\end{equation}
This is a typical example of the cross-geometry situation, as its inputs include both the 2-d manifold functions $\Omega,k_\Omega,f_\Omega$ and the 1-d manifold functions $\partial\Omega,g_{\partial\Omega}$. Since the information provided by $\partial\Omega$ is redundant with that by $\Omega$, here we omit $\partial\Omega$ and only encode $(\Omega, k_\Omega, f_\Omega,g_{\partial\Omega})$, so that the dimension of the initial MFE tokens is 4. The dataset is from \cite{hu2025manifold}, it includes 9000 training examples and 1000 test examples, whose domains $\Omega$ are all contained in $[0,1]^2$. Among these regions, half are disk-like domains and the other half are annulus-like domains, therefore the deformation-based operator learning methods will break down. We compare the MENO with the MFE-based MIONet from \cite{hu2025manifold}, and the results are shown in Table \ref{tab:poisson_cross} and Figure \ref{fig:poisson}. The relative error of MENO is $2.07\%$, one-third of MIONet's $6.49\%$. This case is mainly used to verify the MENO's adaptability to cross-geometry scenarios, while Transolver-3 and PCNO are not applicable.

\begin{table}[H]
  \centering
  \footnotesize
  \setlength{\tabcolsep}{4pt}
  \begin{tabular}{lcccc}
    \toprule
    Model & Rel. error (\%) & Params (M) & Peak GPU mem. (GB) & Train time (h) \\
    \midrule
    MFE-MIONet \cite{hu2025manifold} & 6.49 & 2.54 & 1.16 & 4.70 \\
    \midrule
    MENO (ours) & \textbf{2.07} & 0.721 & 0.912 & 8.40\\
    \bottomrule
  \end{tabular}
  \caption{\textbf{Poisson equation (cross-geometry).} The relative error of MENO is $2.07\%$, one-third of MIONet's $6.49\%$.}
  \label{tab:poisson_cross}
\end{table}

Another dataset of 2-d Poisson equation comes from the PCNO paper \cite{zeng2025point}. In their setting $f\equiv1$ and $g\equiv0$, so that the solution mapping is
\begin{equation}
    \mathcal{G}_{\mathrm{Poisson}}^{\rm single}:\left(\Omega,k_\Omega\right)\mapsto u_\Omega.
\end{equation}
We refer to this as the Poisson equation (single-geometry), to distinguish it from the previous Poisson equation (cross-geometry). The dataset is composed of 1000 training examples, 200 validation examples, and 400 test examples. The results are presented in Table \ref{tab:poisson_single}, which show that MENO achieves slightly higher accuracy than PCNO, while having only $1.7\%$ of the parameters, $5.1\%$ of the peak GPU memory usage, and $8.3\%$ of the training time.

\begin{table}[H]
  \centering
  \footnotesize
  \setlength{\tabcolsep}{4pt}
  \begin{tabular}{lcccc}
    \toprule
    Model & Rel. error (\%) & Params (M) & Peak GPU mem. (GB) & Train time (h) \\
    \midrule
    PCNO \cite{zeng2025point} & 0.683 & 71.5 & 4.67 & 18.2 \\
    \midrule
    MENO (ours) & \textbf{0.669} & 1.27 & 0.239 & 1.52 \\
    \bottomrule
  \end{tabular}
  \caption{\textbf{Poisson equation (single-geometry).} MENO achieves slightly higher accuracy than PCNO, while having only $1.7\%$ of the parameters, $5.1\%$ of the peak GPU memory usage, and $8.3\%$ of the training time.}
  \label{tab:poisson_single}
\end{table}

\subsection{NASA–CRM.}
\begin{figure}[htbp]
    \centering
    \includegraphics[width=.99\textwidth]{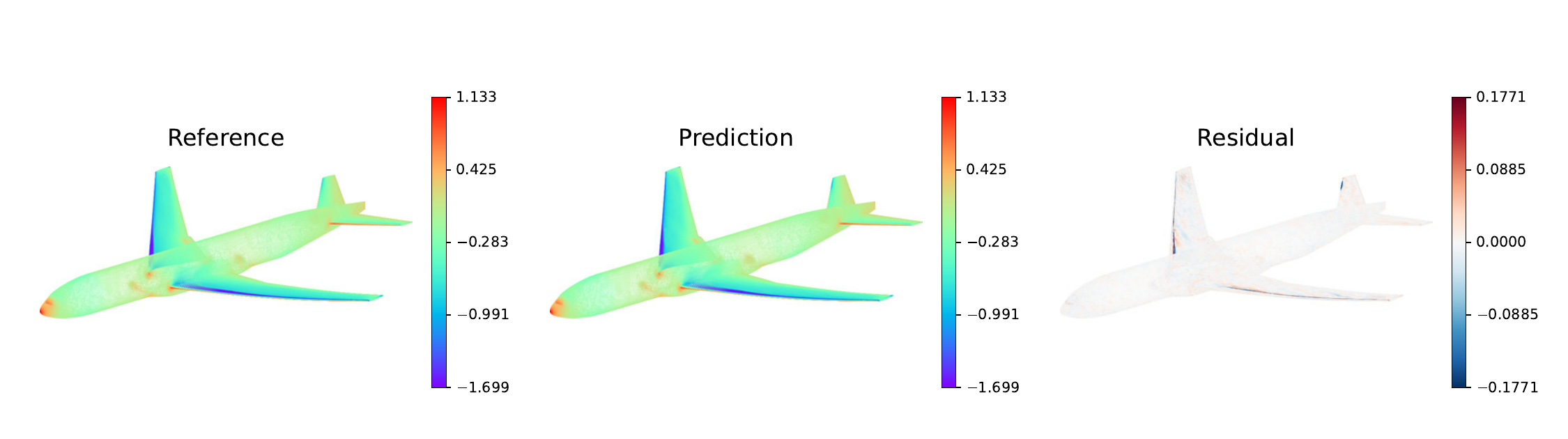}
    \caption{\textbf{Prediction of NASA-CRM.}}
    \label{fig:nasa}
\end{figure}

NASA–CRM is a dataset from \cite{bekemeyer2025introduction}, which is composed of 105 training examples and 44 test examples. This is an aerodynamic problems of aircraft, the solution mapping is
\begin{equation}
    \mathcal{G}_{\mathrm{NASA}}:
\left(\Gamma,\boldsymbol{n}_{\Gamma},
\boldsymbol{\mu}_{\mathrm{NASA}}\right)
\mapsto\left(C_p,\boldsymbol{C}_f\right)\big|_{\Gamma},
\end{equation}
where $\Gamma$ is the surface, $\boldsymbol{n}_{\Gamma}$ is the unit outward normal vector, \(\boldsymbol{\mu}_{\mathrm{NASA}}\in\mathbb{R}^{6}\) contains, in order, the Mach number, angle of attack, inboard aileron deflection, outboard aileron deflection, elevator deflection, and horizontal tailplane setting angle. The outputs \(C_p\) and \(\boldsymbol{C}_f=(C_{f,x},C_{f,y},C_{f,z})\) are the surface pressure coefficient and skin-friction coefficient vector, respectively. The results are shown in Table \ref{tab:nasa} and Figure \ref{fig:nasa}. MENO achieves the errors $6.69\%$ for pressure and $4.79\%$ for friction, better than baselines. All results except for our own and PCNO are cited from the Transolver-3 paper. The hyperparameters of PCNO are identical to those used in the Ahmed body case in the original paper.

\begin{table}[H]
  \centering
  \footnotesize
  \setlength{\tabcolsep}{3pt}
  \begin{tabular}{lccccc}
    \toprule
    Model & Pressure (\%) & Friction (\%) & Params (M) & Peak GPU mem. (GB) & Train time (h) \\
    \midrule
    Graph U-Net \cite{gao2019graph} & 15.85 & 15.61 & n.r. & n.r. & n.r. \\
    GINO \cite{li2023geometry} & 12.39 & 11.51 & n.r. & n.r. & n.r. \\
    MeshGraphNet \cite{pfaff2020learning} & 14.80 & 9.81 & n.r. & n.r. & n.r. \\
    GAOT \cite{wen2026geometry} & 30.38 & 59.79 & n.r. & n.r. & n.r. \\
    UPT \cite{alkin2024universal} & 12.78 & 23.78 & n.r. & n.r. & n.r. \\
    AB-UPT \cite{alkin2025ab} & 9.77 & 6.43 & n.r. & n.r. & n.r. \\
    Transolver-3 \cite{zhou2026transolver} & 8.71 & \underline{5.85} & n.r. & n.r. & n.r. \\
    PCNO \cite{zeng2025point} & \underline{8.20} & 19.5 & 322 & 66.8 & 33.6\\
    \midrule
    MENO (ours) & \textbf{6.69} & \textbf{4.79} & 13.7 & 1.07 & 2.75\\
    \bottomrule
  \end{tabular}
  \caption{\textbf{NASA-CRM.} MENO achieves the errors $6.69\%$ for pressure and $4.79\%$ for friction, better than baselines. All results except for our own and PCNO are cited from the Transolver-3 paper. Here ``n.r.'' means the result is not reported.}
  \label{tab:nasa}
\end{table}

\subsection{AhmedML.}
\begin{figure}[htbp]
    \centering
    \includegraphics[width=.99\textwidth]{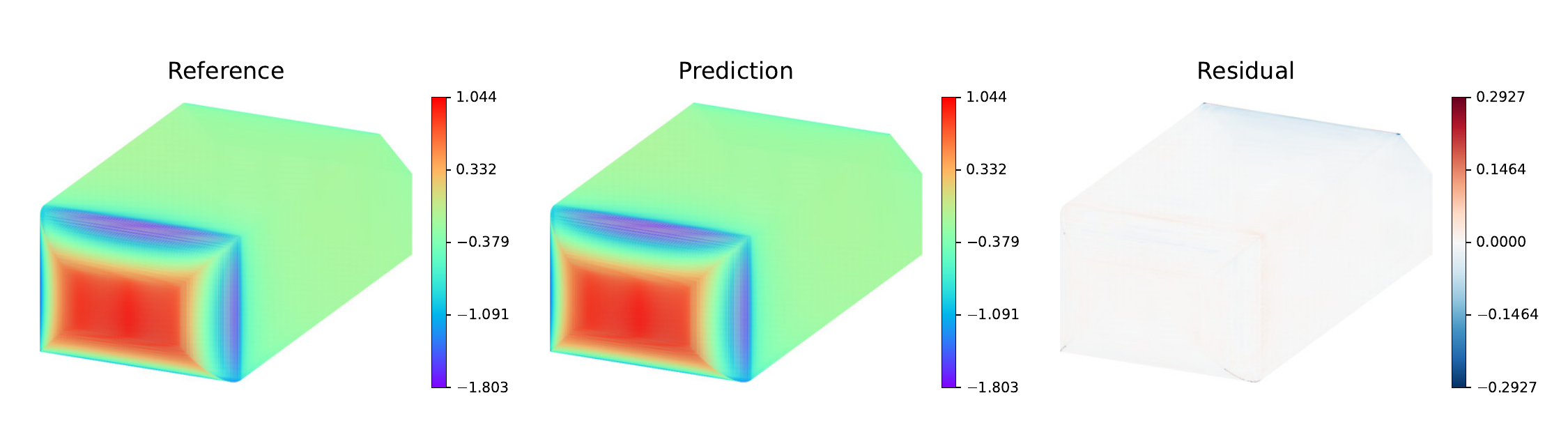}
    \caption{\textbf{Prediction of AhmedML.}}
    \label{fig:ahmedml}
\end{figure}

AhmedML is a dataset from \cite{ashton2024ahmedml}, which is composed of 400 training examples, 50 validation examples and 50 test examples. This is an automotive aerodynamics problem, and the solution mapping is
\begin{equation}
    \mathcal{G}_{\mathrm{Ahmed}}:
\left(\Gamma,\boldsymbol{n}_{\Gamma},
\boldsymbol{\mu}_{\mathrm{Ahmed}}\right)
\mapsto
\left(C_p,\boldsymbol{\tau}_w\right)\big|_{\Gamma},
\end{equation}
where $\Gamma$ is the surface, $\boldsymbol{n}_{\Gamma}$ is the unit outward normal vector, \(\boldsymbol{\mu}_{\mathrm{Ahmed}}\in\mathbb{R}^{8}\) contains, in order, the body length, body height, body width, front-arc diameter, horizontal extent of the rear slant, vertical extent of the rear slant, slant-surface length, and slant angle. The outputs \(C_p\) and \(\boldsymbol{\tau}_w=(\tau_{w,x},\tau_{w,y},\tau_{w,z})\) are the time-averaged surface pressure coefficient and wall-shear-stress vector, respectively. The results are shown in Table \ref{tab:ahmedml} and Figure \ref{fig:ahmedml}. Among all the models, MENO achieves the second-best accuracy. We present another result of small size version of MENO, which also achieves the second-best accuracy, with a very small number of parameters and a very short training time. All results except for our own and PCNO are cited from the Transolver-3 paper. However, the peak memory usage of PCNO exceeds the 80 GB memory limit of a single GPU, resulting in an out-of-memory failure, where the hyperparameters are identical to those used in the Ahmed body case in the original paper. By reducing the number of integration points and decreasing the parameter scale, it is possible to make PCNO work. We will perform more extensive tests in the future.

\begin{table}[H]
  \centering
  \footnotesize
  \setlength{\tabcolsep}{3pt}
  \begin{tabular}{lccccc}
    \toprule
    Model & Pressure (\%) & Friction (\%) & Params (M) & Peak GPU mem. (GB) & Train time (h) \\
    \midrule
    Graph U-Net \cite{gao2019graph} & 6.46 & 7.29  & n.r. & n.r. & n.r. \\
    GINO \cite{li2023geometry} & 7.90 & 8.18 & n.r. & n.r. & n.r. \\
    MeshGraphNet \cite{pfaff2020learning} & 3.72 & 5.01 & n.r. & n.r. & n.r. \\
    GAOT \cite{wen2026geometry} & 8.02 & 9.92  & n.r. & n.r. & n.r. \\
    UPT \cite{alkin2024universal} & 4.25 & 5.80  & n.r. & n.r. & n.r. \\
    AB-UPT \cite{alkin2025ab} & 3.97 & 5.60 & n.r. & n.r. & n.r. \\
    Transolver-3 \cite{zhou2026transolver} & \textbf{2.96} & \textbf{4.81} & n.r. & n.r. & n.r. \\
    PCNO \cite{zeng2025point} & / & / & / & $>80$ & /\\
    \midrule
    MENO (ours, small) & 3.61 & 5.07 & 2.40 & 0.425 & 1.75 \\
    MENO (ours) & \underline{3.37} & \underline{4.88} & 12.1 & 1.94 & 30.5\\
    \bottomrule
  \end{tabular}
  \caption{\textbf{AhmedML.} Among all the models, MENO achieves the second-best accuracy. All results except for our own and PCNO are cited from the Transolver-3 paper. Here ``n.r.'' means the result is not reported, ``/'' indicates not applicable.}
  \label{tab:ahmedml}
\end{table}

\section{Conclusion}
\label{sec:conclusion}
By introducing the Manifold Function Encoder (MFE) for encoding different functions defined on different manifolds, we propose a novel neural operator architecture, the Memory-Efficient Neural Operator (MENO), which has a significantly smaller memory footprint and much faster training speed than other popular architectures, while also exhibiting strong generalization capability. Since the overall architecture of MENO is without any integration operator at the training stage, the GPU memory usage is independent of the data resolution, and this significantly reduces the memory
cost, especially for large-scale problems. MENO can handle PDE inputs of arbitrary form, including arbitrary geometric domains and arbitrary discretizations, and further available for the cross-geometry scenarios. We test several benchmarks, where MENO achieves the best accuracy on Poisson equation and NASA-CRM, and the second-best accuracy on AhmedML. In all cases, the peak GPU memory usage of MENO is significantly lower than that of other architectures.

In future work, we plan to develop general PDE neural solvers based on MENO, owing to its high memory efficiency, cross-geometry handling, and strong generalization capability.

\bibliographystyle{abbrv}
\bibliography{references}

\end{document}